%% file: main.tex
\documentclass[conference]{IEEEtran}
\IEEEoverridecommandlockouts

\usepackage{cite}
\usepackage{amsmath,amssymb,amsfonts,amsthm}
\usepackage{algorithm,algorithmic}
\usepackage{graphicx}
\usepackage{textcomp}
\usepackage{float}
\usepackage{booktabs}
\usepackage[table]{xcolor}
\usepackage{nicefrac}
\usepackage{multirow}
\usepackage[hidelinks]{hyperref}
\usepackage{tikz}
\usepackage{pgfplots}
\usepackage{siunitx}
\usepackage{makecell}

\theoremstyle{plain}
\newtheorem{problem}{Problem}
\newtheorem{theorem}{Theorem}[section]
\newtheorem{proposition}[theorem]{Proposition}
\theoremstyle{definition}
\newtheorem{definition}[theorem]{Definition}
\theoremstyle{remark}
\newtheorem{remark}[theorem]{Remark}

\def\BibTeX{{\rm B\kern-.05em{\sc i\kern-.025em b}\kern-.08em
    T\kern-.1667em\lower.7ex\hbox{E}\kern-.125emX}}

\begin{document}

\title{Leave-One-Out-, Bootstrap- and Cross-\\Conformal Anomaly Detectors
\thanks{This work was conducted as part of the research projects \textit{Biflex Industrie} (grant no. 01MV23020A) and \textit{AutoDiagCM} (grant no. 03EE2046B), funded by the German Federal Ministry for Economic Affairs and Climate Action.}
}

\author{\IEEEauthorblockN{1\textsuperscript{st} Oliver Hennhöfer}
\IEEEauthorblockA{\textit{Intelligent Systems Research Group} \\
\textit{Karlsruhe University of Applied Sciences}\\
Karlsruhe, Germany \\
oliver.hennhoefer@h-ka.de}
\and
\IEEEauthorblockN{2\textsuperscript{nd} Christine Preisach}
\IEEEauthorblockA{\textit{Intelligent Systems Research Group} \\
\textit{Karlsruhe University of Applied Sciences}\\
Karlsruhe, Germany \\
christine.preisach@h-ka.de}
}

\maketitle

\begin{abstract}
The need for uncertainty quantification in anomaly detection systems has become increasingly important. In this context, effectively controlling Type I error rates without inflating Type II error rates in these systems can build trust and reduce costs associated with false discoveries. The field of conformal anomaly detection emerges as a promising approach for providing respective statistical and finite-sample validity guarantees through model calibration. However, reliance on calibration data imposes practical limitations, especially in low-data regimes.\\
In this work, we formally define and evaluate leave-one-out-, bootstrap-, and cross-conformal methods for conformal anomaly detection, building on methods from the field of conformal prediction. Looking beyond the classical split-conformal approach, we show that derived methods for calculating resampling-conformal $p$-values offer a practical compromise between the data efficiency of full-conformal (transductive) approaches and the computational efficiency of split-conformal (inductive) methods. We validate derived methods and quantify their improvements for a range of one-class classifiers and datasets.
\end{abstract}

\begin{IEEEkeywords}
 Conformal Inference, Anomaly Detection, \mbox{Uncertainty Quantification}, False Discovery Rate
\end{IEEEkeywords}

\section{Introduction}

The field of anomaly detection comprises methods for identifying observations that either deviate from the majority or do not \textit{conform} to an expected normal state. Domains of application include cybersecurity \cite{Evangelou2020}, fraud detection \cite{Hilal2022}, predictive maintenance \cite{Carrasco2021, Choi2022}, and healthcare \cite{Fernando2022}, underscoring the relevance of anomaly detection systems in mission-critical industrial applications.

This work focuses on the unsupervised one-class classification approach. This approach is particularly suitable when a representative set of anomalous observations is unavailable, as is the case in most anomaly detection settings.\\
A major limitation shared by one-class classifiers is the lack of statistical guarantees for their estimates. Therefore, an estimator's certainty is by default unquantified, undermining its reliability and trustworthiness. Furthermore, the scarcity of flexible non-parametric models, together with the abundance of parameter-laden algorithms --- both often subject to a priori assumptions and prone to misspecification and overfitting \cite{Keogh2007} --- can result in subpar anomaly estimates and thresholds.

Conformal anomaly detection (CAD) \cite{laxhammar2010, laxhammar2014} seeks to address these problems by leveraging the non-parametric and model-agnostic framework of conformal prediction \cite{Papadopoulos2002, vovk2005, Lei2012} to provide a statistically principled and finite-sample-valid approach to uncertainty quantification.

CAD computes calibrated $p$-values from arbitrary anomaly scores as obtained from any given one-class classifier. The respective $p$-values enable statistical hypothesis testing to determine whether an observation is an inlier \cite{Bates2023}, while controlling the marginal false discovery rate (FDR).

\begin{problem}
    Let $\mathcal{D}$ be a set of observations (inliers) sampled from an arbitrary distribution $P$. Given a new batch of observations $\mathcal{B} = \{x_1, \ldots, x_n\}$, we aim to test the null hypothesis $\mathcal{H}_{0,i}$ for each $x_i \in \mathcal{B}$ as \mbox{$\mathcal{H}_{0,i}: x_i \text{ is drawn from } P \text{ (i.e., is an inlier)}$}. The objective is to determine which observations in $\mathcal{B}$ can be considered as outliers while controlling the marginal FDR for the batch at a specified nominal level $\alpha$.
\end{problem}

The standard split- or inductive conformal procedure splits available (non-anomalous) training data $\mathcal{D}$ into a proper training set $\mathcal{D}_{\text{train}}$ and a calibration set $\mathcal{D}_{\text{calib}}$. After fitting a scoring function $\hat{s}$ with an algorithm $\mathcal{A}$ on $\mathcal{D}_{\text{train}}$, anomaly scores (also conformity scores) are computed on the held-out $\mathcal{D}_{\text{calib}}$. During inference, $p$-values of unseen test observations are then computed as the relative rank of the obtained score among the scores as calculated on $\mathcal{D}_{\text{calib}}$ \cite{Liang2024}.
Resulting statistical guarantees hold when inliers in the calibration and test samples are exchangeable --- a term related to but weaker than the assumption of IID, as it only requires invariance to permutation without independence.

\textbf{Contributions}. Within the given context, the contributions of this work may be summarized as follows:
\begin{itemize}
    \item We formally define leave-one-out-, bootstrap-, and cross-conformal methods for anomaly detection. Respective resampling-conformal methods make more efficient use of available training data than the classical split-conformal (inductive) approach while yielding larger calibration sets, which impact the range of possible $p$-values that can be obtained. Further, we discuss respective theoretical foundations, guarantees, and implications.
    \item We empirically evaluate the marginal FDR and statistical power of the derived methods compared with the split-conformal procedure for Isolation Forest \cite{Liu2008}, $k$-Nearest Neighbors \cite{Cover1967l}, and Isolation-based Anomaly Detection using Nearest-Neighbor Ensembles \cite{Bandaragoda2018} on eight benchmark datasets after the adjustment of obtained $p$-values by the Benjamini--Hochberg procedure \cite{Benjamini1995} for error control.
\end{itemize}

\begin{figure}[H]
    \centering
    \resizebox{0.45\textwidth}{!}{
        \begin{tikzpicture}[font=\rmfamily]
            \fill[transparent] (-8,-8) rectangle (8,8);

            \filldraw[fill=white, draw=black, line width=1pt] (0,0) circle (7.5);
            \node[font=\fontsize{26pt}{16.8pt}\selectfont\bfseries] at (0,5.75) {Conformal Inference};

            \begin{scope}[shift={(-2.8, 0.5)}]
                \filldraw[fill=white, draw=black] (0,0) circle (4.5);
                \node[align=center, font=\fontsize{19pt}{16.8pt}\selectfont\bfseries] at (0,2.6) {Conformal Prediction \\(CP)};
                \filldraw[fill=white, draw=black] (-2.9, 0.5) circle (1.4);
                \node[align=center, font=\fontsize{17pt}{16.8pt}\selectfont\bfseries] at (-2.9, 0.5) {Full CP};
                \filldraw[fill=white, draw=black] (1.2, -0.65) circle (2.4);
                \node[align=center, font=\fontsize{17pt}{16.8pt}\selectfont\bfseries] at (1.2, 0.3) {Inductive CP};
                \filldraw[fill=white, draw=black, line width=1pt] (0.2, -2.4) circle (1.9);
                \node[align=center, font=\fontsize{16pt}{16.8pt}\selectfont\bfseries] at (0.2, -2.4) {Cross-CP \\ LOO-CP \\ Bootstrap-CP};
            \end{scope}

            \begin{scope}[shift={(4.5, -4.4)}]
                \filldraw[fill=black!5, draw=black] (0,0) circle (4);
                \node[align=center, font=\fontsize{17pt}{16.8pt}\selectfont\bfseries] at (0,2) {Conformal Anomaly \\ Detection (CAD)};
                \filldraw[fill=black!5, draw=black] (1.9, -0.35) circle (1.8);
                \node[align=center, font=\fontsize{16pt}{17pt}\selectfont\bfseries] at (1.9, -0.25) {Inductive\\CAD};
                \filldraw[fill=black!10, draw=black, dash pattern=on 4pt off 6pt] (-1, -1.45) circle (2.1);
                \node[align=center, font=\fontsize{16pt}{17pt}\selectfont\bfseries] at (-1, -1.35) {Cross-CAD\\LOO-CAD\\Bootstrap-CAD};
            \end{scope}
        \end{tikzpicture}
    }
    \caption{Non-exhaustive taxonomy of the field of conformal inference with conformal prediction, conformal anomaly detection, and the derived family of \textit{resampling}-conformal methods for anomaly detection.}
    \label{fig:taxonomy}
\end{figure}

\section{Related Work}

Beyond the seminal works regarding conformal inference and conformal prediction \cite{Gammerman1998,Papadopoulos2002,vovk2005,Lei2012}, the term \textit{Conformal Anomaly Detection} was first introduced in \cite{laxhammar2010}.\\
In \cite{laxhammar2010,laxhammar2014}, CAD was initially applied for detecting anomalous trajectories in maritime surveillance applications. The work formalized and discussed the principles of conformal prediction applied to an unsupervised anomaly detection task.

The work of \cite{Bates2023} further advanced the field of CAD by demonstrating that conformal $p$-values are Positive Regression Dependency on a Subset (PRDS) \cite{Benjamini2001} and do not break FDR control via the Benjamini--Hochberg procedure.
The respective work also proposed exploring potentially more powerful variations beyond the inductive approach.

In the context of conformal prediction, the classical inductive approach was initially extended by cross-conformal methods as a ``hybrid of the methods of inductive conformal prediction and cross-validation'' \cite{Vovk2012}, primarily to make more efficient use of available data and induce a higher degree of stability in the calibration procedure. With that, this work mainly builds upon the general idea of ``cross-conformal predictors'' \cite{Vovk2012} and several extensions of the underlying concept, namely Jackknife \cite{Steinberger2016, Steinberger2022}, Jackknife+, CV, CV+ \cite{Vovk2012, Barber2021}, and Jackknife+-after-Bootstrap \cite{Kim2020}.

The underlying idea of leave-one-out and cross-conformal methods for anomaly detection was first formally applied by \cite{Liang2024} for the computation of integrative $p$-values. In this work, one-class classifiers were trained separately on available inliers and outliers to leverage information from both classes and to integrate independently obtained $p$-values into a single scalar statistic. In this context, transductive cross-validation+ (TVC+), based on CV+ \cite{Barber2021}, was proposed. Specifically, the fundamental transductive (also known as full-conformal) approach, owing to its theoretical advantages over the inductive or cross-conformal approaches, was applied.

Other works that apply CAD and related conformal concepts are described in \cite{Smith2015,Guan2022,Haroush2022} and are complemented by \cite{Cai2020,Vovk2021,Vovk2021b}, which focus on the online setting. The works of \cite{Smith2016, Angelopoulos2021, Xu2021, Xu2021b} employed cross-conformal predictors for anomaly detection via a forecasting-based approach.

None of these works explicitly and formally defined, referred to, or empirically evaluated leave-one-out-, bootstrap-, or cross-conformal anomaly detectors.

\section{Background}
\label{sec:background}

Consider a set of data $\mathcal{D}$ comprising $n$ observations $X_i \in \mathbb{R}^d$ in a $d$-dimensional feature space for $i \in [n] = \{1, 2, \dots, n\}$ that were sampled from an unknown continuous, discrete, or mixed distribution $P_X$. The objective is to test the null hypothesis $\mathcal{H}_{0,n+1}$ of whether a new observation $X_{n+1}$ was drawn from $P_X$ under the assumption of exchangeability, i.e., can be considered as an inlier.

\begin{definition}
\label{def:exchangeability}
A sequence $X_{1}, X_{2}, ...X_{n}$ is subject to \textit{exchangeability} when for any permutation $\sigma$ of $\{1, \ldots, n\}$ the joint probability distribution of a permuted sequence $X_{\sigma(1)}, X_{\sigma(2)}, ..., X_{\sigma(n)}$ is identical to the joint probability distribution of the original sequence.
\end{definition}

We aim to compute marginal and superuniform $p$-values $\hat{u}(X_{n+1})$ under
$\mathcal{H}_0$ for all $\alpha \in (0,1)$ with

\begin{equation}
    \mathbb{P}_{\mathcal{H}_{0}}[\hat{u}(X_{n+1}) \le \alpha] \le \alpha,
    \label{eq:1}
\end{equation}

so that $\hat{u}(X_{n+1})$ is super-uniform, see Definition~\ref{def:super1}.

Resulting $p$-values are considered to be marginally
valid as they depend on a subset $\mathcal{D}_{\text{calib}} \subseteq \mathcal{D}$ for calibration and $X_{n+1}$, both are random in Equation~\ref{eq:1}. With that, marginal $p$-values allow for marginal FDR control \cite{Bates2023}.

\subsection{Inductive Conformal Anomaly Detection}
Given $\mathcal{D}$ containing only inliers, the split-conformal approach splits $\mathcal{D}$ into two disjoint subsets $\mathcal{D}_{\text{train}}$ and $\mathcal{D}_{\text{calib}}$.

Data as part of the proper training set $\mathcal{D}_{\text{train}}$ is utilized to fit a one-class classifier to learn a function $\hat{s}(X)$ suitable to compute an anomaly score (or conformity score). In this work, small values of $\hat{s}(X_{n+1})$ are indicative of $X_{n+1}$ being a potential outlier. Consequently, in Equation~\eqref{eq:2}, an outlier receives a small $p$-value as few calibration scores fall at or below its test score. Note that for algorithms where larger scores indicate anomalies, the inequality in Equation~\eqref{eq:2} may be reversed accordingly.

Following the principles of conformal prediction, the score $\hat{s}(X_{n+1})$ of a new observation gets compared to the empirical distribution of $\hat{s}(X_i)$ as computed for the calibration data $\mathcal{D}_{\text{calib}}$, indexed by $i \in \mathcal{D}_{\text{calib}}$. With that, conformal $p$-values are computed as the normalized rank of $\hat{s}(X_{n+1})$ within all $\hat{s}(X_i)$. For any given $X_{n+1} \in \mathbb{R}^d$, the corresponding marginal conformal $p$-value is defined as

\begin{equation}
    \hat{u}(X_{n+1}) = \frac{ \lvert \{X_i \in \mathcal{D}_{\text{calib}} \ : \ \hat{s}(X_{i}) \le \hat{s}(X_{n+1})\}\rvert + 1}{|\mathcal{D}_{\text{calib}}| + 1}
    \label{eq:2}
\end{equation}

The $+1$ correction in the numerator and denominator of Equation~\eqref{eq:2} ensures \textit{super-uniformity}\footnote{\textit{Super-uniformity} is required to validly apply, e.g., the \textit{Benjamini--Hochberg Procedure} \cite{Benjamini1995} for FDR-control, see Section \ref{sec:fdr_control}.} of obtained \textit{p}-values and prevents invalid $p$-values.

\begin{definition}[Super-Uniformity]
\label{def:super1}
A random value $X$ within $[0,1]$ is said to be \textit{super-uniform} if $\mathbb{P}_{\mathcal{H}_{0}}(X \le t) \le t$ for all $t \in [0,1]$. This implies that $X$ is super-uniformly distributed.
\end{definition}

Under the assumption of \textit{exchangeability} the computed conformal \textit{p}-value $\hat{u}(X_{n+1})$ is valid for testing given $\mathcal{H}_{0, n+1}$.

\begin{proposition}[\text{cf. \cite{Bates2023}}]
    \textit{If the inliers in $\mathcal{D}_{\text{calib}}$ are exchangeable with themselves and with $X_{n+1}$, then $\mathbb{P}_{\mathcal{H}_{0}}[\hat{u}(X_{n+1}) \le \alpha] \le \alpha$ for all $\alpha \in (0,1)$}.
    \label{prop:validity_inductive}
\end{proposition}

Besides the marginal validity of computed $p$-values, the range of possible $p$-values to be obtained depends also on $\mathcal{D}_{\text{calib}}$, with the lower bound limited by $\nicefrac{1}{|\mathcal{D}_{\text{calib}}| + 1}$. This poses critical limitations, especially in low-data regimes, as $p$-values might not sufficiently reflect a model's certainty, resulting in conservative yet occasionally anti-conservative $p$-values conditional on the particular realization of $\mathcal{D}_{\text{calib}}$ \cite{Bates2023}. 

Inflated $p$-values solely limited by an insufficiently large $\mathcal{D}_{\text{calib}}$ can prevent powerful downstream statistical error control. This limitation motivates the formal definition and empirical evaluation of leave-one-out-, bootstrap-, and cross-conformal anomaly detectors, which can yield more powerful anomaly detectors in low-data settings by using all available observations in $\mathcal{D}$ for calibration.

\section{Leave-one-out-, Bootstrap- and Cross-Conformal Anomaly Detection}
\label{ccad}

Leave-one-out-, bootstrap-, and cross-conformal anomaly detection extend the
standard split-conformal approach by using resampling schemes for
model training and calibration. Without the need for a dedicated calibration
set \(\mathcal{D}_{\text{calib}}\), respective approaches make more efficient
use of the available data \(\mathcal{D}\), which may be difficult or
expensive to obtain in certain contexts. The resulting anomaly detectors can be less sensitive to unstable estimates arising from unlucky splits, which may induce bias in the calibration procedure. With that, they mitigate sensitivity to a single calibration split.

Throughout this section, we use a conformity-score convention: smaller scores
indicate stronger evidence of anomalousness. If instead \(\hat{s}\) denotes an
anomaly score for which larger values are more anomalous, the inequalities in
the conformal \(p\)-values below have to be reversed.

In principle, the resampling-conformal anomaly detectors construct
subsets of \(\mathcal{D}\) used for model training and calibration. For each
resampling iteration \(k\), a scoring function \(\hat{s}_k\) is learned from
a training subset of \(\mathcal{D}\) by an algorithm \(\mathcal{A}\) suitable
for one-class classification. The observations that were not used to train
\(\hat{s}_k\) are then used to obtain out-of-sample calibration scores.

Algorithm~\ref{alg:rcad} formally defines  \textit{Generalized
Resampling-Conformal Anomaly Detection} that can be parameterized to yield
different variants of conformal anomaly detectors, which will be formally defined in this section.

\begin{algorithm}[h]
\caption{Generalized Resampling-CAD}
\label{alg:rcad}
\begin{algorithmic}[1]
\REQUIRE Inlier data $\mathcal D=\{X_i\}_{i=1}^n$, scoring method $\mathcal A$, test point $X_{n+1}$, resampling scheme $\{T_k\}_{k=1}^K$, aggregation rule $\varphi$, variant $\in\{\mathrm{basic},\mathrm{plus}\}$
\ENSURE Resampling-conformal $p$-value $\hat u(X_{n+1})$

\STATE Let $[n]=\{1,\dots,n\}$.
\STATE For each $k$, let $T_k\subseteq[n]$ denote the training indices and let $H_k=[n]\setminus T_k$ denote the held-out indices.
\STATE For bootstrap resampling, interpret $T_k$ as a multiset and set $H_k=[n]\setminus\operatorname{supp}(T_k)$.
\STATE \textit{Examples:} Jackknife: $T_i=[n]\setminus\{i\}$; $K$-fold CV: $T_k=[n]\setminus S_k$; Bootstrap: $T_k$ is a bootstrap sample and $H_k$ is its out-of-bag set.

\FOR{$k=1,\dots,K$}
    \STATE Fit $\hat s_k \leftarrow \mathcal A(\mathcal D_{T_k})$.
\ENDFOR

\FOR{$i=1,\dots,n$}
    \STATE Let $\mathcal K_i=\{k:i\in H_k\}$ and assume $\mathcal K_i\neq\varnothing$.
    \STATE Define $\hat s_{-i}(x)=\varphi(\{\hat s_k(x):k\in\mathcal K_i\})$.
    \STATE Compute $c_i=\hat s_{-i}(X_i)$.
    \IF{variant $=\mathrm{plus}$}
        \STATE Compute $t_i=\hat s_{-i}(X_{n+1})$.
    \ENDIF
\ENDFOR

\IF{variant $=\mathrm{basic}$}
    \STATE Fit $\hat s\leftarrow\mathcal A(\mathcal D)$ and set $t=\hat s(X_{n+1})$.
    \RETURN $\hat u(X_{n+1})=\bigl(1+\sum_{i=1}^n \mathbf 1\{c_i\le t\}\bigr)/(n+1)$.
\ELSE
    \RETURN $\hat u(X_{n+1})=\bigl(1+\sum_{i=1}^n \mathbf 1\{c_i\le t_i\}\bigr)/(n+1)$.
\ENDIF
\end{algorithmic}
\end{algorithm}

The values returned by Algorithm~\ref{alg:rcad} use up to \(n\) calibration
scores and therefore increase the resolution of the resulting
\(p\)-value grid from
\[
    \frac{1}{|\mathcal{D}_{\text{calib}}|+1}
    \quad\text{to}\quad
    \frac{1}{n+1}.
\]
This increased resolution can enable more powerful detection in downstream error
control, provided $\hat{s}$ adequately separates inliers from
outliers. The resampling procedures considered here follow the practical
cross-conformal construction: models are trained on
subsets of \(\mathcal D=\{X_1,\ldots,X_n\}\), while \(X_{n+1}\) is evaluated
afterward. Hence, the resulting quantities are resampling-conformal
\(p\)-values in the cross-conformal sense, rather than standard conformal \(p\)-values. Their reliability is
therefore assessed through their resampling structure, stability arguments,
and empirical results reported in Section~\ref{sec:evaluation}.

In the following, we formally define leave-one-out-conformal
($\mathrm{J}_{\mathrm{AD}}$, $\mathrm{J}^{+}_{\mathrm{AD}}$),
cross-conformal
($\mathrm{CV}_{\mathrm{AD}}$, $\mathrm{CV}^{+}_{\mathrm{AD}}$),
and bootstrap-conformal
($\mathrm{JaB}_{\mathrm{AD}}$, $\mathrm{J^{+}aB}_{\mathrm{AD}}$)
anomaly detectors.

Throughout, we denote by
\(\mathcal{D}=\{X_1, X_2, \dots, X_n\}\) the full set of available inliers for model training.
When defining leave-one-out and cross-validation subsets, we index into
\(\mathcal{D}\) directly, for example by writing
\(\mathcal{D}\setminus\{X_i\}\), while bootstrap samples \(B_k\) are drawn
from \(\mathcal{D}\) with replacement.

\paragraph{Resampling$^+$ conformal \(p\)-values}

For the resampling$^+$ variants, the conformal score comparison is performed
under the same resampled model for each calibration--test pair. The resulting
\(p\)-value takes the form
\begin{equation}
    \hat{u}(X_{n+1})
    =
    \frac{
    1+
    \sum_{i=1}^n
    \mathbf{1}\bigl(
        \hat{s}_{-i}(X_i)
        \le
        \hat{s}_{-i}(X_{n+1})
    \bigr)
    }{n+1}.
    \label{eq:resampling_plus_pvalue}
\end{equation}

Here, \(\hat{s}_{-i}\) denotes the score function, or aggregated score
function, obtained from resampled models whose training sets exclude \(X_i\).
Thus, each calibration point \(X_i\) and the test point \(X_{n+1}\) are
evaluated under a common score function \(\hat{s}_{-i}\). This pairwise
construction is the central distinction between the resampling$^+$ variants
and their non-retaining counterparts.

\paragraph{Comparison to standard split-conformal validity}

Unlike the split-conformal approach, which partitions the data into a proper training set and a held-out calibration set to achieve exact finite-sample validity, our resampling procedures reuse all \(n\) observations for both training and calibration via different subsets. For each test point \(X_{n+1}\), the scoring models are fit only on subsets of \(\mathcal{D}=\{X_1,\dots,X_n\}\) that exclude individual observations, and \(X_{n+1}\) is evaluated after these models have been fitted. This data reuse aims to gain power by avoiding a dedicated calibration split, but it breaks the independence structure required by the standard conformal rank argument. Consequently, the resulting resampling-conformal \(p\)-values are heuristic approximations to split-conformal \(p\)-values. Their reliability is assessed through algorithmic stability considerations and empirical evidence rather than guaranteed by a finite-sample proof.

\paragraph{Dependence and multiple testing}

The conformity scores for calibration
\[
    \hat{s}_{-1}(X_1), \dots, \hat{s}_{-n}(X_n)
\]
are computed from potentially overlapping training sets and are therefore dependent. This
dependence is part of the resampling design and has to be accounted for when the resulting \(p\)-values are used in multiple testing. In particular, FDR
control via the Benjamini--Hochberg procedure relies on suitable dependence
conditions, such as PRDS, that are discussed in
Section~\ref{sec:fdr_control}.

\begin{remark}[Non-retaining variants and stability]
For the non-retaining variants introduced in the following
(\(\mathrm{J}_{\mathrm{AD}}\), \(\mathrm{CV}_{\mathrm{AD}}\),
\(\mathrm{JaB}_{\mathrm{AD}}\)), the score for a test point $N_{n+1}$ is computed from
\(\hat{s}\) trained on all of \(\mathcal{D}\), while calibration scores use leave-one-out, cross-validation, or bootstrap-based score functions
that excludes the corresponding calibration point. Their reliability is
therefore closely tied to algorithmic stability: when
\(\hat{s}\) and the corresponding resampled score functions are uniformly
close, the non-retaining and resampling$^+$ variants yield similar
decisions \cite{Barber2021}. This motivates the use of the retaining
variants when inference costs are acceptable. In particular, while the retaining variants are not exactly conformal in the present
resampling setting, they remain theoretically preferable because they do
not require this additional full-data-versus-resampled centering
approximation.
\label{rem:stability}
\end{remark}

\subsection{$\mathrm{Jackknife}_{\mathrm{AD}}$ and $\mathrm{Jackknife}^{+}_{\mathrm{AD}}$}
\label{jackknife}

The term Jackknife \cite{Quenouille1949, Quenouille1956, Tukey1958} denotes a statistical procedure encompassing general resampling techniques for estimating bias and variance of a statistical estimator \cite{Shao1995}. In contrast, the well-known leave-one-out validation can be viewed as a specific implementation of Jackknife for model evaluation in machine learning.

Following the Jackknife procedure, we define $n$ leave-one-out sets $\mathcal{D} \setminus \{X_i\}$ for $i \in \{1,\dots,n\}$. For the standard $\mathrm{Jackknife}_{\mathrm{AD}}$ ($\mathrm{J}_{\mathrm{AD}}$), described in \cite{Barber2021,Steinberger2016,Steinberger2022} for predictive tasks, we fit a scoring function

\begin{equation}
    \hat{s}_{-i} := \mathcal{A}(\mathcal{D} \setminus \{X_i\})
\end{equation}

and compare each resulting calibration score $\hat{s}_{-i}(X_i)$ to the test score $\hat{s}(X_{n+1})$ as obtained by a function $\hat{s}$, subsequently fitted on all available $X_i \in \mathcal{D}$ as

\begin{equation}
    \hat{u}(X_{n+1}) =
    \frac{
        \sum_{i=1}^n
        \mathbf{1}\left(\hat{s}_{-i}(X_i) \le \hat{s}(X_{n+1})\right) + 1
    }{n + 1}.
    \label{eq:jackknife}
\end{equation}

For the $\mathrm{Jackknife}^{+}_{\mathrm{AD}}$ ($\mathrm{J}^{+}_{\mathrm{AD}}$), we additionally retain all $n$ fitted scoring functions $\hat{s}_{-i}$. To compute the conformity score, each calibration score is paired with the test score generated by the same leave-one-out model, maintaining statistical symmetry as

\begin{equation}
    \hat{u}(X_{n+1}) =
    \frac{
        \sum_{i=1}^n
        \mathbf{1}\left(\hat{s}_{-i}(X_i) \le \hat{s}_{-i}(X_{n+1})\right) + 1
    }{n + 1}.
    \label{eq:jackknife_plus}
\end{equation}

The $\mathrm{J}^{+}_{\mathrm{AD}}$ is an extension to $\mathrm{J}_{\mathrm{AD}}$ primarily seeking to stabilize obtained anomaly estimates by centering them around the corresponding leave-one-out estimates $\hat{s}_{-i}(X_{n+1})$ instead of a single $\hat{s}(X_{n+1})$. As long as the estimator is not highly sensitive to certain observations (or subsets) of $\mathcal{D}$, the results are similar \cite{Barber2021}. The \(\mathrm{J}^{+}_{\mathrm{AD}}\) construction improves comparability by evaluating
each calibration point and the test point under the same leave-one-out model.
However, as discussed above, this pairwise symmetry does not by itself yield
an exchangeable score vector over \(X_1,\dots,X_n,X_{n+1}\). Thus,
\(\mathrm{J}^{+}_{\mathrm{AD}}\) should be interpreted as a resampling-based
approximation to conformal anomaly detection unless additional assumptions,
such as suitable algorithmic stability, are imposed, whereas
$\mathrm{J}_{\mathrm{AD}}$ relies on $\hat{s} \approx \hat{s}_{-i}$
(Remark~\ref{rem:stability}).

Both methods become expensive on larger datasets, and $\mathrm{J+}_{\mathrm{AD}}$ additionally incurs substantial inference-time cost because all leave-one-out models must be retained and evaluated.

\subsection{$\mathrm{CV}_{\mathrm{AD}}$ and $\mathrm{CV}^{+}_{\mathrm{AD}}$}
\label{subsec:cv}

$\mathrm{CV}_{\mathrm{AD}}$ and $\mathrm{CV}^{+}_{\mathrm{AD}}$ can be seen as generalizations of $\mathrm{J}_{\mathrm{AD}}$ and $\mathrm{J}^{+}_{\mathrm{AD}}$ by creating $K$ disjoint folds $S_1, S_2, \dots, S_K$\footnote{For simplicity, we assume $K$ divides $n$. In practice, fold sizes may differ by at most one observation.}, with $\bigcup_{k=1}^{K} S_k = \mathcal{D}$, fitting $K$ scoring functions

\begin{equation}
    \hat{s}_{-S_k} := \mathcal{A}(\mathcal{D} \setminus S_k),
\end{equation}

to calculate $\mathrm{CV}_{\mathrm{AD}}$ in analogy to $\mathrm{J}_{\mathrm{AD}}$ as

\begin{equation}
    \hat{u}(X_{n+1}) =
    \frac{
        \sum_{k=1}^K \sum_{X_i \in S_k}
        \mathbf{1}\left(\hat{s}_{-S_k}(X_i) \le \hat{s}(X_{n+1})\right) + 1
    }{n + 1}
\end{equation}

and $\mathrm{CV}^{+}_{\mathrm{AD}}$, respectively, in analogy to $\mathrm{J}^{+}_{\mathrm{AD}}$ as

\begin{equation}
    \hat{u}(X_{n+1}) =
    \frac{
        \sum_{k=1}^K \sum_{X_i \in S_k}
        \mathbf{1}\left(\hat{s}_{-S_k}(X_i) \le \hat{s}_{-S_k}(X_{n+1})\right) + 1
    }{n + 1}.
\end{equation}

The advantage of $\mathrm{CV}_{\mathrm{AD}}$ and $\mathrm{CV}^{+}_{\mathrm{AD}}$ is naturally the lower computational cost, depending on the parameterization of $K$. Compared to the leave-one-out case, using $K < n$ folds introduces a pessimistic bias, as each scoring function is fit on a smaller training set, but may reduce the variance of the resulting resampling-based estimates, as the per-fold fits are less correlated.

\subsection{Jackknife- and Jackknife$^{+}$-after-Bootstrap$_{\mathrm{AD}}$}
\label{jackknife_after_bootstrap}

The originally conceived Jackknife$^+$-after-Bootstrap ($\mathrm{J^{+}aB}_{\mathrm{AD}}$) \cite{Kim2020} is based on the idea of Jackknife$^+$-after-Bootstrap \cite{efron1992} and may analogously be extended by the respective non-retaining variant. The procedure iteratively draws $K$ bootstrap samples $B_1, B_2, \dots, B_K$ of size $n$ with replacement from $\mathcal{D}$ to fit $K$ scoring functions:

\begin{equation}
    \hat{s}_{B_k} := \mathcal{A}(B_k).
\end{equation}

To compute the conformity scores, the method evaluates each instance $X_i \in \mathcal{D}$ strictly on the subset of models for which $X_i$ was out-of-bag. Let $C_i = \{k \in \{1, \ldots, K\} \mid X_i \notin B_k\}$ denote the index set of these models\footnote{We require $C_i \neq \emptyset$ for all $i$. Any observation with no out-of-bag models is excluded from the calibration set. Since each observation is out-of-bag with probability $(1-\nicefrac{1}{n})^n \approx e^{-1}$ per sample, $\mathbb{P}(C_i = \emptyset) = (1 - (1-\nicefrac{1}{n})^n)^K$ vanishes rapidly with $K$ (e.g., ${<}\,10^{-4}$ for $K \geq 25$).}. Using an aggregation function $\varphi(\cdot)$ (e.g., $\operatorname{Median}[\cdot]$), the aggregated out-of-bag scoring function for $X_i$ is defined as:

\begin{equation}
    \hat{s}_{-i}(x) := \varphi\left(\{ \hat{s}_{B_k}(x) \}_{k \in C_i}\right).
\end{equation}

For the standard, non-retaining $\mathrm{JaB}_{\mathrm{AD}}$, these $n$ out-of-bag scores are compared against a test score obtained from a final model $\hat{s}$ trained on all available data $\mathcal{D}$ as

\begin{equation}
    \hat{u}(X_{n+1}) =
    \frac{
        \sum_{i=1}^n \mathbf{1}\left(\hat{s}_{-i}(X_i) \le \hat{s}(X_{n+1})\right) + 1
    }{n + 1}.
\end{equation}

For the model-retaining variant $\mathrm{J^{+}aB}_{\mathrm{AD}}$, each out-of-bag calibration score is instead evaluated against the test score generated by its corresponding aggregated out-of-bag model $\hat{s}_{-i}$, establishing model calibration and inference as

\begin{equation}
    \hat{u}(X_{n+1}) =
    \frac{
        \sum_{i=1}^n \mathbf{1}\left(\hat{s}_{-i}(X_i) \le \hat{s}_{-i}(X_{n+1})\right) + 1
    }{n + 1}.
\end{equation}

Note that bootstrap sampling with replacement introduces duplicate observations within each $B_k$, meaning that individual bootstrap training sets are not strictly exchangeable draws from $P_X$. However, the out-of-bag mechanism ensures that each calibration score $\hat{s}_{-i}(X_i)$ is computed exclusively from models that did not include $X_i$ during training, and the aggregation via $\varphi(\cdot)$ reduces the resulting set of scores to a single score per observation. For $\mathrm{J^{+}aB}_{\mathrm{AD}}$, the out-of-bag construction ensures that each calibration point is evaluated only by models that did not train on that point, and the corresponding test score is evaluated using the same aggregated out-of-bag model. This gives the same pairwise comparability as in the other resampling$^+$ variants. However, because the resampling scheme is still applied only to $\mathcal{D}$ and not to the augmented sample $\mathcal{D} \cup \{X_{n+1}\}$, this does not by itself imply exact finite-sample conformal validity. For $\mathrm{JaB}_{\mathrm{AD}}$, the same stability caveat as for the other non-retaining variants applies (see Remark~\ref{rem:stability}). This construction mirrors the Jackknife$^{+}$-after-Bootstrap procedure of~\cite{Kim2020}, originally established for conformal prediction with predictive intervals.

\input{results/knn}
\input{results/iforest}

\section{Multiple Testing, PRDS, and the Benjamini-Hochberg Procedure}
\label{sec:fdr_control}

In order to control the FDR, the $p$-values used in the Benjamini--Hochberg procedure should be marginally super-uniform and satisfy suitable dependence conditions. In the split-conformal setting, marginal validity follows from the usual conformal argument because the calibration scores are computed on a held-out set independent of the test point. By contrast, our resampling variants reuse data via overlapping training and calibration splits, so we do not have a strict finite-sample validity guarantee. Any theoretical guarantee would require additional stability assumptions that we do not discuss here. Accordingly, the resampling methods are treated as approximate, and their error rates are evaluated empirically.

The FDR is defined as the expected proportion of false discoveries ($Q$), given by the ratio of erroneous discoveries ($V$) to total discoveries ($R$), with $\mathbb{E}(Q)=0$ if $R=0$:

\begin{equation}
    \mathrm{FDR} = \mathbb{E}(Q) = \mathbb{E}\left[\frac{V}{\max(R, 1)}\right].
\end{equation}

The Benjamini--Hochberg procedure computes adjusted $p$-values $p^{\mathrm{BH}}_{(i)}$ for $m$ tested hypotheses $\{\mathcal{H}_{0i}\}_{i=1}^{m}$ with corresponding $p$-values $\{p_i\}_{i=1}^{m}$ sorted as $p_{(1)} \le p_{(2)} \le \dots \le p_{(m)}$. Let $k$ be the largest index such that $p_{(k)} \le \nicefrac{k\alpha}{m}$. If no such $k$ exists, no discovery is made. Otherwise, reject the $k$ hypotheses corresponding to $p_{(1)}, \dots, p_{(k)}$, with adjusted $p$-values defined as

\begin{equation}
    p^{\mathrm{BH}}_{(i)} = \min\left(\min_{j \ge i} \frac{m \, p_{(j)}}{j}, 1\right).
\end{equation}

For the split-conformal setting, \cite{Bates2023} proved that conformal $p$-values are PRDS on the set of true nulls, establishing BH validity. The key mechanism is that all test-point $p$-values are rank-based functions of a shared calibration set: larger calibration scores simultaneously decrease all $p$-values, and vice versa, inducing the required monotonic dependence.

\begin{definition}[\textbf{PRDS}, e.g.\ from~\cite{Benjamini2001}]
A vector of $p$-values $p = (p_1, \dots, p_m)$ is \emph{PRDS on} $I_0 \subseteq \{1, \dots, m\}$ if for every $i \in I_0$ and every non-decreasing measurable set $D \subseteq [0, 1]^m$, the map $t \mapsto \mathbb{P}(p \in D \mid p_i \le t)$ is non-decreasing in $t \in [0, 1]$.
\end{definition}

\begin{theorem}[BH under PRDS, e.g.\ from~\cite{Benjamini2001}]
If the null $p$-values are super-uniform and the joint distribution of $(p_1, \dots, p_m)$ is PRDS on the subset corresponding to true null hypotheses, then the BH procedure controls the FDR at level at most $\nicefrac{m_0}{m}\alpha$, where $m_0$ is the number of true nulls.
\end{theorem}

\paragraph{Extension to resampling-conformal $p$-values}

In the resampling-conformal setting, the calibration scores $\{\hat{s}_{-i}(X_i)\}_{i=1}^n$ are computed from models trained on overlapping subsets of $\mathcal{D}$, introducing dependencies among them that are absent in the split-conformal case. The monotonic coupling that yields PRDS for split-conformal CAD arises because all test-point $p$-values are functions of the same calibration set. In our resampling-based methods, the calibration scores come from overlapping subsets, so it is unclear whether the same positive dependence holds. Establishing PRDS in this setting would require new proofs and is beyond the scope of this work. We therefore do not claim unconditional PRDS for the resampling-conformal $p$-values.

We note that conditional on the calibration scores $\{\hat{s}_{-i}(X_i)\}_{i=1}^n$, each test-point $p$-value $\hat{u}(X_j^{\mathrm{test}})$ depends deterministically only on the corresponding test score $\hat{s}(X_j^{\mathrm{test}})$ (or $\hat{s}_{-i}(X_j^{\mathrm{test}})$ for the $+$ variants). Conditional on the fitted models and calibration scores, the test points are independent, and the resulting $p$-values are deterministic functions of the corresponding test scores. This conditional structure is suggestive of the split-conformal PRDS mechanism, but it does not by itself establish unconditional PRDS for the resampling-based procedures. A formal proof would need to account for the dependence induced by overlapping training sets and shared fitted models.

\section{Evaluation}
\label{sec:evaluation}

We evaluate the split-, leave-one-out-, cross- ($K = 10$), and bootstrap-conformal ($K = 100$) methods across several benchmark datasets to assess their effectiveness in providing reliable uncertainty quantification in anomaly detection. Respective empirical results should be interpreted as evidence for the practical reliability of the proposed methods, not as a finite-sample guarantee of FDR control.

The conformal methods are applied to \textit{Isolation Forest} \cite{Liu2008}, \textit{k-Nearest Neighbors} (KNN), and \textit{Isolation-based Anomaly Detection using Nearest-Neighbor Ensembles} (INNE) \cite{Bandaragoda2018}. The algorithms are used with their default hyperparameters as implemented by \texttt{PyOD} \cite{Zhao2019}.

Eight benchmark datasets from the \mbox{\textit{ADBench}} benchmark collection \cite{Han2022} are used for evaluation. The datasets are selected to encompass diverse data in terms of size and dimensionality (see Table~\ref{tab:data}).

Due to their computational costs, $\mathrm{Jackknife}_{\mathrm{AD}}$ and $\mathrm{Jackknife}^{+}_{\mathrm{AD}}$ are only evaluated on datasets with $\mathcal{D}_{\mathrm{train}} < 500$. The FDR is controlled at $\alpha = 0.1$ via the BH procedure.

\begin{table}[!htbp]
    \centering
    \caption{Evaluation datasets as part of \textit{ADBench} \cite{Han2022}.}
    \label{tab:data}

    \begin{tabular}{llrr}
        \toprule
        & Name
        & $n$
        & $n_{\mathrm{feature}}$ \\
        \midrule

        \multirow{4}{*}{\rotatebox[origin=c]{90}{\footnotesize Low Data}}
        & \textbf{WBC}          & 223    & 9   \\
        & \textbf{Ionosphere}   & 351    & 33  \\
        & \textbf{WDBC}         & 367    & 30  \\
        & \textbf{Breast}       & 683    & 9   \\

        \addlinespace

        \multirow{4}{*}{\rotatebox[origin=c]{90}{\footnotesize High Data}}
        & \textbf{Musk}         & 3,062  & 166 \\
        & \textbf{Thyroid}      & 3,772  & 6   \\
        & \textbf{Satellite}    & 6,435  & 36  \\
        & \textbf{Mammography}  & 11,183 & 6   \\

        \bottomrule
    \end{tabular}
\end{table}

\input{results/inne}

\subsection{Experimental Setup}
\label{subsec:setup}

Following the general setup as described in \cite{Bates2023}, we randomly draw $J = 25$ distinct datasets $\mathcal{D}_{1}, \dots, \mathcal{D}_{J}$, comprising only \textit{normal} observations. Each dataset $\mathcal{D}_{j}$ is independent and used for training and calibration. In the implemented procedure, each $\mathcal{D}_{j}$ is paired with $L = 100$ randomly drawn test sets $\mathcal{D}_{j,l}^{\mathrm{test}}$, each containing both normal observations and outliers disjoint from $\mathcal{D}_j$.

For the evaluation, we are interested in the FDR conditional on the realised $\mathcal{D}_{j}$, defined as

\begin{equation}
    \mathrm{cFDR}(\mathcal{D}_{j})
    :=
    \mathbb{E}\left[
        \mathrm{FDP}\left(\mathcal{D}^{\mathrm{test}}; \mathcal{D}_{j}\right)
        \mid \mathcal{D}_{j}
    \right],
\end{equation}

where $\mathrm{FDP}\left(\mathcal{D}^{\mathrm{test}}; \mathcal{D}_{j}\right)$ denotes the false discovery proportion, i.e., the proportion of reported outliers that are in fact normal observations. If no outliers are reported, the FDP is zero.

The results for any given $j \in \{1, \dots, J\}$ are evaluated by

\begin{equation}
    \widehat{\mathrm{cFDR}}(\mathcal{D}_{j})
    :=
    \frac{1}{L}
    \sum_{l=1}^{L}
    \mathrm{FDP}\left(\mathcal{D}_{j,l}^{\mathrm{test}}; \mathcal{D}_{j}\right),
\end{equation}

and the statistical power is evaluated by

\begin{equation}
    \widehat{\mathrm{cPower}}(\mathcal{D}_{j})
    :=
    \frac{1}{L}
    \sum_{l=1}^{L}
    \mathrm{Power}\left(\mathcal{D}_{j,l}^{\mathrm{test}}; \mathcal{D}_{j}\right),
\end{equation}

where $\mathrm{Power}\left(\mathcal{D}_{j,l}^{\mathrm{test}}; \mathcal{D}_{j}\right)$ is defined as the proportion of total outliers in $\mathcal{D}_{j,l}^{\mathrm{test}}$ correctly identified as outliers.

Our experiments seek to evaluate whether the \textit{marginal} FDR is controlled, computed as

\begin{equation}
    \widehat{\mathrm{mFDR}}
    =
    \frac{1}{J}
    \sum_{j=1}^{J}
    \widehat{\mathrm{cFDR}}(\mathcal{D}_{j}).
\end{equation}

The nominal validity requirement is

\begin{equation}
    \mathrm{mFDR}
    :=
    \mathbb{E}_{\mathcal{D}}\left[\mathrm{cFDR}(\mathcal{D})\right]
    \leq q,
\end{equation}

where $q$ denotes the target FDR level. Since $\widehat{\mathrm{mFDR}}$ is estimated from a finite number of randomized datasets, small numerical exceedances of $q$ are not interpreted as validity violations unless they are large relative to the implied Monte Carlo variability. Let

\[
    X_j := \widehat{\mathrm{cFDR}}(\mathcal{D}_j),
    \qquad
    s_X^2
    :=
    \frac{1}{J - 1}
    \sum_{j=1}^{J}
    \left(X_j - \widehat{\mathrm{mFDR}}\right)^2.
\]

We regard a method as showing empirical evidence of a validity violation only if the one-sided lower confidence bound

\begin{equation}
    \widehat{\mathrm{mFDR}}
    -
    t_{0.95, J - 1}
    \frac{s_X}{\sqrt{J}}
\end{equation}

exceeds $q$. Otherwise, the observed value is considered statistically compatible with the expected marginal FDR control.

\subsection{Implementation Details}
\label{subsec:details}

For training and calibration, $n_{\mathrm{inlier}}/2$ observations were used ($n_{\mathrm{train}} = \lfloor n_{\mathrm{inlier}}/2 \rfloor$). For each dataset $\mathcal{D}_{j}$, the conformal method was fit and calibrated once, and the resulting procedure was evaluated on all $L = 100$ associated test sets. For the split-conformal approach, these observations were partitioned into $\mathcal{D}_{\mathrm{train}}$ and $\mathcal{D}_{\mathrm{calib}}$, each of approximately the size $n_{\mathrm{train}}/2$. The resampling-conformal methods used all $n_{\mathrm{train}}$ observations for both training and calibration as described in Section~\ref{ccad}. This reflects the intended trade-off: given a fixed data budget, the resampling methods make more efficient use of available data by avoiding a dedicated calibration split. We control the anomaly rate at $\lambda \approx 0.1$ to ensure sufficient discoveries for meaningful comparisons across all methods.

\section{Results and Discussion}

The results in Table~\ref{fig:result_knn}, Table~\ref{fig:result_iforest}, and Table~\ref{fig:result_inne} extend the empirical findings of \cite{Bates2023} by showing that resampling-conformal $p$-values can exhibit reliable FDR behavior in a marginal sense in our experiments with target level $\mathrm{mFDR} \leq 0.1$ ($\alpha = 0.1$). The single exceedance observed for the stricter retaining variants does not constitute empirical evidence of a systematic violation under the present Monte Carlo design.

With regard to statistical power, the resampling-conformal methods empirically outperform the split-conformal approach. This can mainly be attributed to the larger calibration sets produced by the resampling-based methods, which yield lower $p$-values that remain significant after multiple testing correction. As the size of the calibration set has diminishing marginal benefit, the advantage of resampling-conformal methods diminishes in high-data regimes.

Looking at the bigger picture, several factors are decisive for the performance of the conformal methods. First, the usefulness of obtained \textit{conformity scores} (and resulting $p$-values) is primarily determined by the learned scoring function \cite{Angelopoulos2021}. An unsuitable algorithm $\mathcal{A}$ learning a deficient scoring function $\hat{s}$ will fail to produce powerful $p$-values. Further, violations of the exchangeability assumption between calibration and test data may affect the validity of the resulting $p$-values. Lastly, evaluating the impact of retaining trained classifiers via $\mathrm{J}^{+}_{\mathrm{AD}}$, $\mathrm{CV}^{+}_{\mathrm{AD}}$ and $\mathrm{J^{+}aB}_{\mathrm{AD}}$ reveals a clear practical trade-off. While neither the retaining nor the non-retaining variants inherit exact finite-sample validity guarantees in the present resampling setting, the retaining variants remain theoretically preferable because each calibration--test comparison is carried out under the same resampled scoring function. The non-retaining variants can nevertheless be attractive in practice and may exhibit higher statistical power. Their main advantage is practical: they avoid the need to retain and evaluate the fitted classifiers at inference time, making them substantially easier to deploy in production. However, the evaluation reveals that FDR control violations, while rare, can unexpectedly spike, as in the case for \texttt{INNE} with $\mathrm{JaB}_{\mathrm{AD}}$ on the \texttt{Musk} dataset.

The key findings of our evaluation may be summarized as:
\begin{itemize}
    \item Resampling$^+$-based methods in CAD offer reliable empirical control of the marginal FDR via the Benjamini--Hochberg procedure and tend to yield more powerful anomaly detectors, particularly for fewer training data.
    \item Calibration set sizes have a decreasing marginal benefit, as the advantages of resampling- over split-conformal methods become smaller within \textit{high}-data regimes.
    \item When approximate error control suffices, the non-retaining variants offer a viable alternative for production deployments under algorithmic stability assumptions.
\end{itemize}

\paragraph*{Remarks}
All results should be interpreted within the general multiple-testing setting, with the Benjamini--Hochberg procedure used here as a representative method for error control. In downstream applications, the effectiveness of CAD also depends strongly on the test-batch size, the chosen nominal level $\alpha$, and the anomaly rate. Larger batches, smaller values of $\alpha$, and lower anomaly rates require smaller $p$-values for discoveries to be made. Accordingly, the benefit of resampling-based methods in low-data regimes is relative: it becomes particularly relevant when more conservative multiple-testing settings make lower attainable $p$-values increasingly important.

\section{Conclusion}

Resampling-based methods represent a natural and practically useful extension of CAD. They are particularly promising in low-data regimes, where efficient use of available inlier data is important. In our experiments, the resulting resampling$^+$-conformal $p$-values generally outperformed classical split-conformal $p$-values without exhibiting systematic violations of error control validity, despite their weaker theoretical support.

By framing anomaly detection as a multiple-testing problem, the marginal FDR of resampling-CAD can be assessed empirically using the Benjamini--Hochberg procedure, while the proposed methods often exhibit higher statistical power. Due to the model-agnostic nature of conformal methods, the proposed procedures can be integrated easily into existing anomaly detection pipelines. Their main limitations are the increased computational cost, especially during training, and their general theoretical reliance on exchangeability.

In addition, CAD integrates naturally with anomaly detection algorithms such as \textit{Isolation Forest}, which otherwise require a manually chosen threshold or contamination estimate. Overall, our results support the usefulness of general conformal ideas for detecting conformal anomalies beyond the classical split-conformal case. More broadly, this work formally defines leave-one-out-, bootstrap-, and cross-conformal anomaly detectors as a unified family of resampling-conformal methods for conformal anomaly detection.

\section*{Reproducibility}

Conducted experiments are accessible under \href{https://github.com/OliverHennhoefer/resampling-cad}{\texttt{github.com/OliverHennhoefer/resampling-cad}} for exact reproduction (via \texttt{Python}). Applied conformal methods are implemented in our publicly available \mbox{\texttt{PyPI} package} \href{https://github.com/OliverHennhoefer/nonconform}{\texttt{nonconform}} (\texttt{Python}).

\bibliographystyle{IEEEtran}
\bibliography{references}

\end{document}

%% file: results/knn.tex
\begin{table*}[h]
    \caption{Performance of split- and resampling-conformal methods using $k$-Nearest Neighbors. Marginal FDR ($\alpha=0.1$) and mean statistical power ($\bar{x}$) are reported as mean $\pm$ standard deviation. \textbf{Bold} marks the highest valid power within the retaining ($+$) and non-retaining blocks. No marginal FDR exceedances were observed.}
    \begin{tabular}{llllllll}
        & \multicolumn{7}{c}{\textbf{$k$-Nearest Neighbors}} \\ \toprule
        & \multicolumn{7}{c}{False Discovery Rate} \\ \cmidrule{2-8}
        & \multicolumn{1}{c}{}
        & \multicolumn{3}{c}{\textbf{Retaining ($+$)}}
        & \multicolumn{3}{c}{\textbf{Non-Retaining}} \\
        \cmidrule(lr){3-5} \cmidrule(lr){6-8}
        & \multicolumn{1}{c}{Split}
        & \multicolumn{1}{c}{JaB$^{+}_{\mathrm{AD}}$}
        & \multicolumn{1}{c}{10-CV$^{+}_{\mathrm{AD}}$}
        & \multicolumn{1}{c}{Jackknife$^{+}_{\mathrm{AD}}$}
        & \multicolumn{1}{c}{JaB$_{\mathrm{AD}}$}
        & \multicolumn{1}{c}{10-CV$_{\mathrm{AD}}$}
        & \multicolumn{1}{c}{Jackknife$_{\mathrm{AD}}$} \\
        \cmidrule(lr){2-2} \cmidrule(lr){3-5} \cmidrule(lr){6-8}
        WBC
        & 0.034 $\pm$ 0.077
        & 0.078 $\pm$ 0.095
        & 0.058 $\pm$ 0.078
        & 0.061 $\pm$ 0.084
        & 0.077 $\pm$ 0.095
        & 0.053 $\pm$ 0.073
        & 0.061 $\pm$ 0.084 \\
        \rowcolor{lightgray!30} Ionosphere
        & 0.003 $\pm$ 0.007
        & 0.047 $\pm$ 0.080
        & 0.061 $\pm$ 0.101
        & 0.061 $\pm$ 0.096
        & 0.046 $\pm$ 0.080
        & 0.049 $\pm$ 0.088
        & 0.061 $\pm$ 0.096 \\
        WDBC
        & 0.076 $\pm$ 0.083
        & 0.092 $\pm$ 0.088
        & 0.084 $\pm$ 0.077
        & 0.087 $\pm$ 0.081
        & 0.090 $\pm$ 0.084
        & 0.083 $\pm$ 0.076
        & 0.087 $\pm$ 0.081 \\
        \rowcolor{lightgray!30} Breast
        & 0.046 $\pm$ 0.050
        & 0.081 $\pm$ 0.078
        & 0.076 $\pm$ 0.079
        & 0.080 $\pm$ 0.073
        & 0.078 $\pm$ 0.075
        & 0.073 $\pm$ 0.077
        & 0.080 $\pm$ 0.073 \\
        \addlinespace
        Musk
        & 0.082 $\pm$ 0.037
        & 0.104 $\pm$ 0.034
        & 0.049 $\pm$ 0.023
        & ---
        & 0.093 $\pm$ 0.036
        & 0.044 $\pm$ 0.021
        & --- \\
        \rowcolor{lightgray!30} Thyroid
        & 0.032 $\pm$ 0.054
        & 0.073 $\pm$ 0.088
        & 0.073 $\pm$ 0.089
        & ---
        & 0.073 $\pm$ 0.088
        & 0.071 $\pm$ 0.087
        & --- \\
        Satellite
        & 0.078 $\pm$ 0.042
        & 0.082 $\pm$ 0.040
        & 0.078 $\pm$ 0.039
        & ---
        & 0.081 $\pm$ 0.040
        & 0.075 $\pm$ 0.038
        & --- \\
        \rowcolor{lightgray!30} Mammography
        & 0.046 $\pm$ 0.061
        & 0.064 $\pm$ 0.050
        & 0.061 $\pm$ 0.051
        & ---
        & 0.063 $\pm$ 0.048
        & 0.058 $\pm$ 0.050
        & --- \\
        \toprule
        & \multicolumn{7}{c}{Statistical Power} \\ \cmidrule{2-8}
        & \multicolumn{1}{c}{}
        & \multicolumn{3}{c}{\textbf{Retaining ($+$)}}
        & \multicolumn{3}{c}{\textbf{Non-Retaining}} \\
        \cmidrule(lr){3-5} \cmidrule(lr){6-8}
        & \multicolumn{1}{c}{Split}
        & \multicolumn{1}{c}{JaB$^{+}_{\mathrm{AD}}$}
        & \multicolumn{1}{c}{10-CV$^{+}_{\mathrm{AD}}$}
        & \multicolumn{1}{c}{Jackknife$^{+}_{\mathrm{AD}}$}
        & \multicolumn{1}{c}{JaB$_{\mathrm{AD}}$}
        & \multicolumn{1}{c}{10-CV$_{\mathrm{AD}}$}
        & \multicolumn{1}{c}{Jackknife$_{\mathrm{AD}}$} \\
        \cmidrule(lr){2-2} \cmidrule(lr){3-5} \cmidrule(lr){6-8}
        WBC
        & 0.072 $\pm$ 0.159
        & \textbf{0.638 $\pm$ 0.192}
        & 0.586 $\pm$ 0.218
        & 0.604 $\pm$ 0.193
        & \textbf{0.648 $\pm$ 0.185}
        & 0.580 $\pm$ 0.213
        & 0.604 $\pm$ 0.193 \\
        \rowcolor{lightgray!30} Ionosphere
        & 0.006 $\pm$ 0.013
        & 0.248 $\pm$ 0.149
        & 0.259 $\pm$ 0.177
        & \textbf{0.275 $\pm$ 0.166}
        & 0.244 $\pm$ 0.151
        & 0.245 $\pm$ 0.166
        & \textbf{0.275 $\pm$ 0.166} \\
        WDBC
        & 0.382 $\pm$ 0.339
        & \textbf{0.782 $\pm$ 0.258}
        & 0.771 $\pm$ 0.262
        & 0.772 $\pm$ 0.263
        & \textbf{0.782 $\pm$ 0.258}
        & 0.771 $\pm$ 0.262
        & 0.772 $\pm$ 0.263 \\
        \rowcolor{lightgray!30} Breast
        & 0.347 $\pm$ 0.216
        & \textbf{0.770 $\pm$ 0.076}
        & 0.739 $\pm$ 0.102
        & 0.766 $\pm$ 0.077
        & 0.764 $\pm$ 0.078
        & 0.726 $\pm$ 0.102
        & \textbf{0.766 $\pm$ 0.077} \\
        \addlinespace
        Musk
        & \textbf{1.000 $\pm$ 0.000}
        & \textbf{1.000 $\pm$ 0.000}
        & \textbf{1.000 $\pm$ 0.000}
        & ---
        & \textbf{1.000 $\pm$ 0.000}
        & \textbf{1.000 $\pm$ 0.000}
        & --- \\
        \rowcolor{lightgray!30} Thyroid
        & 0.018 $\pm$ 0.033
        & \textbf{0.066 $\pm$ 0.038}
        & 0.061 $\pm$ 0.029
        & ---
        & \textbf{0.065 $\pm$ 0.038}
        & 0.060 $\pm$ 0.029
        & --- \\
        Satellite
        & 0.354 $\pm$ 0.016
        & \textbf{0.367 $\pm$ 0.016}
        & 0.363 $\pm$ 0.013
        & ---
        & \textbf{0.367 $\pm$ 0.016}
        & 0.362 $\pm$ 0.013
        & --- \\
        \rowcolor{lightgray!30} Mammography
        & 0.041 $\pm$ 0.067
        & \textbf{0.049 $\pm$ 0.044}
        & 0.043 $\pm$ 0.048
        & ---
        & \textbf{0.048 $\pm$ 0.044}
        & 0.042 $\pm$ 0.047
        & --- \\
        \bottomrule
    \end{tabular}
    \label{fig:result_knn}
\end{table*}

%% file: results/iforest.tex
\begin{table*}[h]
    \caption{Performance of split- and resampling-conformal methods using Isolation Forest. Marginal FDR ($\alpha=0.1$) and mean statistical power ($\bar{x}$) are reported as mean $\pm$ standard deviation. \textbf{Bold} marks the highest valid power within the retaining ($+$) and non-retaining blocks. No marginal FDR exceedances were observed.}
    \begin{tabular}{llllllll}
        & \multicolumn{7}{c}{\textbf{Isolation Forest}} \\ \toprule
        & \multicolumn{7}{c}{False Discovery Rate} \\ \cmidrule{2-8}
        & \multicolumn{1}{c}{}
        & \multicolumn{3}{c}{\textbf{Retaining ($+$)}}
        & \multicolumn{3}{c}{\textbf{Non-Retaining}} \\
        \cmidrule(lr){3-5} \cmidrule(lr){6-8}
        & \multicolumn{1}{c}{Split}
        & \multicolumn{1}{c}{JaB$^{+}_{\mathrm{AD}}$}
        & \multicolumn{1}{c}{10-CV$^{+}_{\mathrm{AD}}$}
        & \multicolumn{1}{c}{Jackknife$^{+}_{\mathrm{AD}}$}
        & \multicolumn{1}{c}{JaB$_{\mathrm{AD}}$}
        & \multicolumn{1}{c}{10-CV$_{\mathrm{AD}}$}
        & \multicolumn{1}{c}{Jackknife$_{\mathrm{AD}}$} \\
        \cmidrule(lr){2-2} \cmidrule(lr){3-5} \cmidrule(lr){6-8}
        WBC
        & 0.013 $\pm$ 0.047
        & 0.066 $\pm$ 0.079
        & 0.077 $\pm$ 0.086
        & 0.076 $\pm$ 0.091
        & 0.089 $\pm$ 0.089
        & 0.066 $\pm$ 0.083
        & 0.076 $\pm$ 0.091 \\
        \rowcolor{lightgray!30} Ionosphere
        & 0.001 $\pm$ 0.003
        & 0.016 $\pm$ 0.039
        & 0.009 $\pm$ 0.026
        & 0.010 $\pm$ 0.029
        & 0.018 $\pm$ 0.049
        & 0.011 $\pm$ 0.026
        & 0.010 $\pm$ 0.029 \\
        WDBC
        & 0.064 $\pm$ 0.101
        & 0.055 $\pm$ 0.078
        & 0.057 $\pm$ 0.096
        & 0.087 $\pm$ 0.098
        & 0.054 $\pm$ 0.085
        & 0.052 $\pm$ 0.094
        & 0.090 $\pm$ 0.101 \\
        \rowcolor{lightgray!30} Breast
        & 0.041 $\pm$ 0.056
        & 0.081 $\pm$ 0.075
        & 0.088 $\pm$ 0.083
        & 0.093 $\pm$ 0.082
        & 0.084 $\pm$ 0.074
        & 0.086 $\pm$ 0.083
        & 0.093 $\pm$ 0.082 \\
        \addlinespace
        Musk
        & 0.035 $\pm$ 0.071
        & 0.012 $\pm$ 0.028
        & 0.003 $\pm$ 0.012
        & ---
        & 0.065 $\pm$ 0.088
        & 0.022 $\pm$ 0.053
        & --- \\
        \rowcolor{lightgray!30} Thyroid
        & 0.067 $\pm$ 0.049
        & 0.067 $\pm$ 0.038
        & 0.067 $\pm$ 0.040
        & ---
        & 0.069 $\pm$ 0.041
        & 0.082 $\pm$ 0.046
        & --- \\
        Satellite
        & 0.085 $\pm$ 0.046
        & 0.096 $\pm$ 0.030
        & 0.084 $\pm$ 0.031
        & ---
        & 0.100 $\pm$ 0.048
        & 0.096 $\pm$ 0.051
        & --- \\
        \rowcolor{lightgray!30} Mammography
        & 0.048 $\pm$ 0.045
        & 0.075 $\pm$ 0.060
        & 0.054 $\pm$ 0.058
        & ---
        & 0.068 $\pm$ 0.065
        & 0.072 $\pm$ 0.067
        & --- \\
        \toprule
        & \multicolumn{7}{c}{Statistical Power} \\ \cmidrule{2-8}
        & \multicolumn{1}{c}{}
        & \multicolumn{3}{c}{\textbf{Retaining ($+$)}}
        & \multicolumn{3}{c}{\textbf{Non-Retaining}} \\
        \cmidrule(lr){3-5} \cmidrule(lr){6-8}
        & \multicolumn{1}{c}{Split}
        & \multicolumn{1}{c}{JaB$^{+}_{\mathrm{AD}}$}
        & \multicolumn{1}{c}{10-CV$^{+}_{\mathrm{AD}}$}
        & \multicolumn{1}{c}{Jackknife$^{+}_{\mathrm{AD}}$}
        & \multicolumn{1}{c}{JaB$_{\mathrm{AD}}$}
        & \multicolumn{1}{c}{10-CV$_{\mathrm{AD}}$}
        & \multicolumn{1}{c}{Jackknife$_{\mathrm{AD}}$} \\
        \cmidrule(lr){2-2} \cmidrule(lr){3-5} \cmidrule(lr){6-8}
        WBC
        & 0.027 $\pm$ 0.098
        & 0.345 $\pm$ 0.313
        & 0.412 $\pm$ 0.342
        & \textbf{0.442 $\pm$ 0.338}
        & \textbf{0.517 $\pm$ 0.336}
        & 0.398 $\pm$ 0.338
        & 0.442 $\pm$ 0.338 \\
        \rowcolor{lightgray!30} Ionosphere
        & 0.001 $\pm$ 0.005
        & 0.023 $\pm$ 0.058
        & \textbf{0.016 $\pm$ 0.040}
        & 0.018 $\pm$ 0.048
        & \textbf{0.030 $\pm$ 0.075}
        & 0.018 $\pm$ 0.042
        & 0.018 $\pm$ 0.048 \\
        WDBC
        & 0.232 $\pm$ 0.372
        & 0.188 $\pm$ 0.290
        & 0.220 $\pm$ 0.362
        & \textbf{0.325 $\pm$ 0.392}
        & 0.207 $\pm$ 0.341
        & 0.190 $\pm$ 0.361
        & \textbf{0.330 $\pm$ 0.395} \\
        \rowcolor{lightgray!30} Breast
        & 0.220 $\pm$ 0.260
        & 0.523 $\pm$ 0.228
        & 0.561 $\pm$ 0.224
        & \textbf{0.574 $\pm$ 0.244}
        & \textbf{0.592 $\pm$ 0.229}
        & 0.566 $\pm$ 0.231
        & 0.574 $\pm$ 0.244 \\
        \addlinespace
        Musk
        & 0.081 $\pm$ 0.152
        & \textbf{0.102 $\pm$ 0.270}
        & 0.003 $\pm$ 0.008
        & ---
        & \textbf{0.197 $\pm$ 0.323}
        & 0.097 $\pm$ 0.230
        & --- \\
        \rowcolor{lightgray!30} Thyroid
        & 0.544 $\pm$ 0.307
        & 0.569 $\pm$ 0.266
        & \textbf{0.582 $\pm$ 0.271}
        & ---
        & 0.579 $\pm$ 0.279
        & \textbf{0.645 $\pm$ 0.234}
        & --- \\
        Satellite
        & 0.394 $\pm$ 0.047
        & \textbf{0.400 $\pm$ 0.035}
        & 0.389 $\pm$ 0.029
        & ---
        & \textbf{0.395 $\pm$ 0.057}
        & 0.391 $\pm$ 0.055
        & --- \\
        \rowcolor{lightgray!30} Mammography
        & 0.036 $\pm$ 0.028
        & \textbf{0.058 $\pm$ 0.037}
        & 0.049 $\pm$ 0.031
        & ---
        & 0.057 $\pm$ 0.055
        & \textbf{0.060 $\pm$ 0.063}
        & --- \\
        \bottomrule
    \end{tabular}
    \label{fig:result_iforest}
\end{table*}

%% file: results/inne.tex
\begin{table*}[h]
    \caption{Performance of split- and resampling-conformal methods using Isolation-based Nearest-Neighbor Ensemble. Marginal FDR ($\alpha=0.1$) and mean statistical power ($\bar{x}$) are reported as mean $\pm$ standard deviation. \textbf{Bold} marks the highest valid power within the retaining ($+$) and non-retaining blocks. FDR exceedances are marked by \textsuperscript{\dag}.}
    \begin{tabular}{llllllll}
        & \multicolumn{7}{c}{\textbf{Isolation-based Nearest-Neighbor Ensemble}} \\ \toprule
        & \multicolumn{7}{c}{False Discovery Rate} \\ \cmidrule{2-8}
        & \multicolumn{1}{c}{}
        & \multicolumn{3}{c}{\textbf{Retaining ($+$)}}
        & \multicolumn{3}{c}{\textbf{Non-Retaining}} \\
        \cmidrule(lr){3-5} \cmidrule(lr){6-8}
        & \multicolumn{1}{c}{Split}
        & \multicolumn{1}{c}{JaB$^{+}_{\mathrm{AD}}$}
        & \multicolumn{1}{c}{10-CV$^{+}_{\mathrm{AD}}$}
        & \multicolumn{1}{c}{Jackknife$^{+}_{\mathrm{AD}}$}
        & \multicolumn{1}{c}{JaB$_{\mathrm{AD}}$}
        & \multicolumn{1}{c}{10-CV$_{\mathrm{AD}}$}
        & \multicolumn{1}{c}{Jackknife$_{\mathrm{AD}}$} \\
        \cmidrule(lr){2-2} \cmidrule(lr){3-5} \cmidrule(lr){6-8}
        WBC
        & 0.004 $\pm$ 0.013
        & 0.026 $\pm$ 0.056
        & 0.028 $\pm$ 0.075
        & 0.013 $\pm$ 0.042
        & 0.017 $\pm$ 0.045
        & 0.022 $\pm$ 0.059
        & 0.013 $\pm$ 0.040 \\
        \rowcolor{lightgray!30} Ionosphere
        & 0.019 $\pm$ 0.049
        & 0.057 $\pm$ 0.067
        & 0.052 $\pm$ 0.065
        & 0.044 $\pm$ 0.054
        & 0.055 $\pm$ 0.069
        & 0.050 $\pm$ 0.066
        & 0.052 $\pm$ 0.067 \\
        WDBC
        & 0.077 $\pm$ 0.081
        & 0.085 $\pm$ 0.081
        & 0.075 $\pm$ 0.079
        & 0.081 $\pm$ 0.080
        & 0.088 $\pm$ 0.082
        & 0.076 $\pm$ 0.073
        & 0.080 $\pm$ 0.078 \\
        \rowcolor{lightgray!30} Breast
        & 0.031 $\pm$ 0.062
        & 0.049 $\pm$ 0.081
        & 0.034 $\pm$ 0.075
        & 0.021 $\pm$ 0.047
        & 0.041 $\pm$ 0.072
        & 0.032 $\pm$ 0.071
        & 0.024 $\pm$ 0.054 \\
        \addlinespace
        Musk
        & 0.086 $\pm$ 0.041
        & 0.076 $\pm$ 0.026
        & 0.020 $\pm$ 0.019
        & ---
        & \textbf{0.189 $\pm$ 0.102\textsuperscript{\dag}}
        & 0.101 $\pm$ 0.094
        & --- \\
        \rowcolor{lightgray!30} Thyroid
        & 0.066 $\pm$ 0.050
        & 0.092 $\pm$ 0.041
        & 0.067 $\pm$ 0.047
        & ---
        & 0.100 $\pm$ 0.046
        & 0.066 $\pm$ 0.046
        & --- \\
        Satellite
        & 0.071 $\pm$ 0.039
        & 0.084 $\pm$ 0.040
        & 0.077 $\pm$ 0.031
        & ---
        & 0.082 $\pm$ 0.043
        & 0.077 $\pm$ 0.036
        & --- \\
        \rowcolor{lightgray!30} Mammography
        & 0.061 $\pm$ 0.068
        & 0.104 $\pm$ 0.050
        & 0.064 $\pm$ 0.051
        & ---
        & 0.112 $\pm$ 0.050
        & 0.073 $\pm$ 0.064
        & --- \\
        \toprule
        & \multicolumn{7}{c}{Statistical Power} \\ \cmidrule{2-8}
        & \multicolumn{1}{c}{}
        & \multicolumn{3}{c}{\textbf{Retaining ($+$)}}
        & \multicolumn{3}{c}{\textbf{Non-Retaining}} \\
        \cmidrule(lr){3-5} \cmidrule(lr){6-8}
        & \multicolumn{1}{c}{Split}
        & \multicolumn{1}{c}{JaB$^{+}_{\mathrm{AD}}$}
        & \multicolumn{1}{c}{10-CV$^{+}_{\mathrm{AD}}$}
        & \multicolumn{1}{c}{Jackknife$^{+}_{\mathrm{AD}}$}
        & \multicolumn{1}{c}{JaB$_{\mathrm{AD}}$}
        & \multicolumn{1}{c}{10-CV$_{\mathrm{AD}}$}
        & \multicolumn{1}{c}{Jackknife$_{\mathrm{AD}}$} \\
        \cmidrule(lr){2-2} \cmidrule(lr){3-5} \cmidrule(lr){6-8}
        WBC
        & 0.006 $\pm$ 0.014
        & 0.051 $\pm$ 0.120
        & \textbf{0.063 $\pm$ 0.186}
        & 0.028 $\pm$ 0.101
        & 0.044 $\pm$ 0.122
        & \textbf{0.045 $\pm$ 0.137}
        & 0.033 $\pm$ 0.122 \\
        \rowcolor{lightgray!30} Ionosphere
        & 0.032 $\pm$ 0.079
        & \textbf{0.275 $\pm$ 0.155}
        & 0.228 $\pm$ 0.169
        & 0.239 $\pm$ 0.150
        & \textbf{0.263 $\pm$ 0.172}
        & 0.229 $\pm$ 0.169
        & 0.246 $\pm$ 0.161 \\
        WDBC
        & 0.393 $\pm$ 0.313
        & \textbf{0.803 $\pm$ 0.244}
        & 0.718 $\pm$ 0.338
        & 0.776 $\pm$ 0.262
        & \textbf{0.767 $\pm$ 0.307}
        & 0.726 $\pm$ 0.343
        & 0.740 $\pm$ 0.323 \\
        \rowcolor{lightgray!30} Breast
        & 0.075 $\pm$ 0.170
        & \textbf{0.122 $\pm$ 0.175}
        & 0.069 $\pm$ 0.124
        & 0.051 $\pm$ 0.090
        & \textbf{0.108 $\pm$ 0.160}
        & 0.067 $\pm$ 0.121
        & 0.059 $\pm$ 0.107 \\
        \addlinespace
        Musk
        & \textbf{1.000 $\pm$ 0.000}
        & \textbf{1.000 $\pm$ 0.000}
        & \textbf{1.000 $\pm$ 0.000}
        & ---
        & 1.000 $\pm$ 0.000
        & \textbf{1.000 $\pm$ 0.000}
        & --- \\
        \rowcolor{lightgray!30} Thyroid
        & 0.234 $\pm$ 0.181
        & \textbf{0.302 $\pm$ 0.098}
        & 0.208 $\pm$ 0.128
        & ---
        & \textbf{0.303 $\pm$ 0.118}
        & 0.235 $\pm$ 0.150
        & --- \\
        Satellite
        & 0.344 $\pm$ 0.012
        & \textbf{0.351 $\pm$ 0.008}
        & 0.350 $\pm$ 0.008
        & ---
        & \textbf{0.352 $\pm$ 0.010}
        & 0.350 $\pm$ 0.010
        & --- \\
        \rowcolor{lightgray!30} Mammography
        & 0.051 $\pm$ 0.059
        & \textbf{0.068 $\pm$ 0.013}
        & 0.042 $\pm$ 0.023
        & ---
        & \textbf{0.095 $\pm$ 0.050}
        & 0.057 $\pm$ 0.044
        & --- \\
        \bottomrule
    \end{tabular}
    \label{fig:result_inne}
\end{table*}